%% file: main.tex
\documentclass{article}

\usepackage{arxiv}
\input{math_commands.tex}

\usepackage[table,xcdraw]{xcolor}

\usepackage{subfigure}
\usepackage{nicefrac}       
\usepackage{microtype}      
\usepackage{lipsum}

\usepackage{picinpar,graphicx}
\usepackage{hyperref}
\usepackage{url}
\usepackage{amsmath,bm,amssymb,latexsym}
\usepackage{graphicx}
\usepackage{subfigure}
\usepackage{threeparttable}

\usepackage{times}
\usepackage{epsfig}
\usepackage{graphicx}
\usepackage{amsmath}
\usepackage{amssymb}
\usepackage{times}
\usepackage{epsfig}

\usepackage{mathrsfs}
\usepackage{amsthm}
\usepackage{setspace}
\usepackage{booktabs}
\usepackage{amsfonts}
\usepackage{subfigure}
\usepackage{multirow}
\usepackage{float}
\usepackage{subeqnarray}
\usepackage{cases}
\usepackage{makeidx}
\usepackage{float}
\usepackage{colortbl}
\usepackage{natbib}

\title{ Interactive Model with Structural Loss for Language-based Abductive Reasoning }

\author{
\\
\bigskip
\bf Linhao Li\textsuperscript{1},
Ming Xu\textsuperscript{1},
Yongfeng Dong\textsuperscript{1},
Xin Li\textsuperscript{2},
Ao Wang\textsuperscript{1}
\\
{1} The School of Artificial Intellgence and the Hebei Provincial Key Laboratory of Big Data Computing, \\Hebei University of Technology
\\
{2} The Lane Department of Computer Science and Electrical Engineering,\\West Virginia University
\\
\bigskip
\\
\{lilinhao,dongyf\}@hebut.edu.cn,
201932804008@stu.hebut.edu.cn,
\\
Xin.Li@mail.wvu.edu,202132803143@stu.hebut.edu.cn
}
\renewcommand{\undertitle}{Under consideration at Pattern Recognition Letters}
\begin{document}
\maketitle
\begin{abstract}

The abductive natural language inference task ($\alpha$NLI) is proposed to infer the most plausible explanation between the cause and the event. In the $\alpha$NLI task, two observations are given, and the most plausible hypothesis is asked to pick out from the candidates. Existing methods model the relation between each candidate hypothesis separately and penalize the inference network uniformly. In this paper, we argue that it is unnecessary to distinguish the reasoning abilities among correct hypotheses; and similarly, all wrong hypotheses contribute the same when explaining the reasons of the observations. Therefore, we propose to group instead of ranking the hypotheses and design a structural loss called ``joint softmax focal loss'' in this paper. Based on the observation that the hypotheses are generally semantically related, we have designed a novel interactive language model aiming at exploiting the rich interaction among competing hypotheses. We name this new model for $\alpha$NLI: Interactive Model with Structural Loss (IMSL). The experimental results show that our IMSL has achieved the highest performance on the RoBERTa-large pretrained model, with ACC and AUC results increased by about 1\% and  5\% respectively.
\end{abstract}

\section{Introduction}

Abductive natural language inference ($\alpha$NLI) \citep{DBLP:conf/iclr/BhagavatulaBMSH20} is a newly established branch of natural language inference (NLI) and is an interesting task in the area of natural language processing (NLP) based commonsense reasoning. Originating from NLI which targets at the semantic relationship between the two sentences, $\alpha$NLI further estimates the abductive reasoning of each sentence by explicitly deducing its cause. In the past years, $\alpha$NLI has attracted increasing attentions as it makes NLP tools more explainable and comprehensible. As of today, typical applications of $\alpha$NLI include knowledge graph Completion \citep{DBLP:conf/akbc/YuZSNS20} \citep{DBLP:conf/eacl/BauerB21}, question answering \citep{DBLP:conf/aaai/MaIFBNO21}, sentence in-filling \citep{DBLP:conf/acl/HuangZEC20}, knowledge integration \citep{DBLP:conf/iclr/ZhouLSL021} and so on.

To better motivate this work, we have shown a comparison between NLI and $\alpha$NLI in Table \ref{table:task}. For NLI, the task is to judge the relationship between the premise statement $\rm{P}$ and the hypothetical sentence $\rm{H}$ based on the given information in $\rm{P}$. Options of the answer can be implication, neutrality, or contradiction. For $\alpha$NLI, a pair of observations ($\rm O_1$ and $\rm O_2$) and some hypotheses (e.g., two competing hypotheses $\rm H^1$ and $\rm H^2$ in the example) are given. The task of $\alpha$NLI is to deduce the more plausible reason between $\rm H^1$ and $\rm H^2$ that can explain the situational change from $\rm O_1$ to $\rm O_2$. In addition to constructing the $\alpha$NLI task, the authors of \cite{DBLP:conf/iclr/BhagavatulaBMSH20} has released a new challenge data set, called ART and reported comprehensive baseline performance for $\alpha$NLI by directly employing and retraining a solution for NLI, i.e., ESIM+ELMo\citep{DBLP:conf/acl/ChenZLWJI17,DBLP:conf/naacl/PetersNIGCLZ18}. They also found that the pretrained language model can apparently influence the performance of an algorithm and demonstrated some test results with the latest language models like GPT\citep{radford2018improving} and BERT\citep{DBLP:conf/naacl/DevlinCLT19}. 
\begin{table}[htb]
\begin{center}
\caption{Comparison of NLI tasks and $\alpha$NLI tasks, where E, N, and C represent entailment, neutral and contradiction, respectively} \label{table:task}
\begin{tabular}{|l|l|c|}
\hline
\rowcolor[HTML]{D0CECE} 
Task                                         & Context                                                                & \multicolumn{1}{l|}{\cellcolor[HTML]{D0CECE}Answer} \\ \hline
                                             & P:  A man inspects the uniform of a figure in some East Asian country. &                                                     \\
                                             & \quad H: The man is sleeping.                                                & \multirow{-2}{*}{E , N or \textbf{C}}                        \\
                                             & P: An older and younger man smiling.                                   &                                                     \\
                                             & \quad H: Two men are smiling and laughing at the cats playing on the floor.  & \multirow{-2}{*}{E , \textbf{N} or C}                        \\
                                             & P: A soccer game with multiple males playing.                          &                                                     \\
\multirow{-6}{*}{NLI}                        & \quad H: Some men are playing a sport.                                       & \multirow{-2}{*}{\textbf{E} , N or {C}}                        \\ \hline
\multicolumn{1}{|c|}{}                       & $\rm O_1$: Dotty was being very grumpy.                                       &                                                     \\
\multicolumn{1}{|c|}{}                       & \quad $\rm H^1$: Dotty ate something bad.                                           &                                                     \\
\multicolumn{1}{|c|}{}                       & \quad $\rm H^2$: Dotty call some close friends to chat.                             &                                                     \\
\multicolumn{1}{|c|}{\multirow{-4}{*}{$\alpha$NLI}} & $\rm O_2$: She felt much better afterwards.                                   & \multirow{-4}{*}{$\rm{H^1}$ or $ \mathbf {H^2}$}                          \\ \hline
\end{tabular}
\end{center}
\end{table}

We note that there is still a considerable gap between the human performance and the class of baseline models in \cite{DBLP:conf/iclr/BhagavatulaBMSH20}. More recently, \cite{DBLP:conf/sigir/ZhuPLC20} argued that the former framework cannot measure the rationality of the hypotheses, and reformulated $\alpha$NLI as a learning-to-rank task for abductive reasoning. In their approach, RoBERTa\citep{DBLP:journals/corr/abs-1907-11692}, BERT\citep{DBLP:conf/naacl/DevlinCLT19}, and ESIM\citep{DBLP:conf/acl/ChenZLWJI17} are all tested to work as the pretrained language model. Under this new ranking-based framework, \cite{DBLP:conf/emnlp/PaulF20} introduces a novel multi-head knowledge attention model which learns to focus on multiple pieces of knowledge at the same time, and is capable of refining the input representation in a recursive manner for $\alpha$NLI.

\begin{figure}[htb]
\begin{center}
\includegraphics[width=140mm]{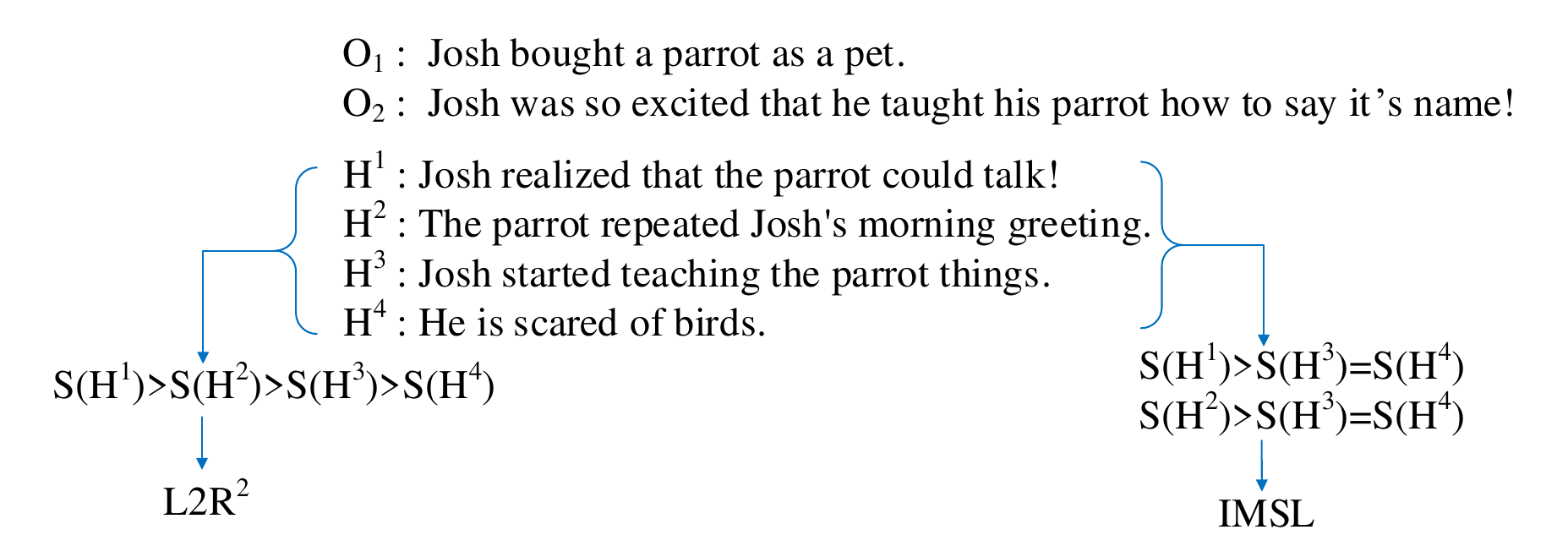}
\end{center}
\vspace{-0.2in}
\caption{Comparison of $\rm L2R^2$ method and IMSL method. Among them, $\rm O_1$, $\rm O_2$ represent observations, $\rm H^1$, $\rm H^2$ are correct answers, $\rm H^3$, $\rm H^4$ are wrong answers, S($\rm H^i$) represents the score of the i-th hypothesis correctness.} \label{fig:method}
\end{figure}

Despite the performance improvement achieved by the ranking framework, there  are still some weaknesses calling for further investigation. For instance, a practical training sample (e.g., two observations and four hypotheses) from ART is shown in Figure \ref{fig:method}. It is easy to conclude that both $\rm H^1$ and $\rm H^2$ are correct answers; while the other two ($\rm H^3, \rm H^4$) are false. However, in previous ranking-based $\alpha$NLI method such as $\rm L2R^2$ (Learning to Rank for Reasoning) (Zhu et al., 2020, four hypotheses will be trained simultaneously by treating one of the two correct answers as a more correct one. Similarly, the wrong answers are also treated as a wrong one and a worse one. Meanwhile, the ranked hypotheses are trained separately, but the sum of their probabilities is set as a fixed value - e.g., the probability of correct hypothesis $\rm H^2$ decreases when the probability of answer $\rm H^1$ increases.

In this paper, we advocate a new approach  for $\alpha$NLI as shown in Figure \ref{fig:method}. Our principle of abductive reasoning is constructed based on following two arguments: 1) a hypothesis is correct because its meaning explains the change of the observations. In practice, the causes of situational changes are often diverse, and therefore the answers are seldom unique. It follows that we  do not need to intentionally distinguish or rank the correct answers. 2) a hypothesis is wrong because it can not explain the cause of some event. Therefore, all wrong answers contribute the same - i.e., it is plausible to treat all wrong hypotheses equally in the process of constructing our reasoning network. We argue that the proposed abductive reasoning principle is closer to commonsense reasoning by humans than previous ranking-based approaches.

\begin{figure}[htb]
\begin{center}
\includegraphics[width=105mm]{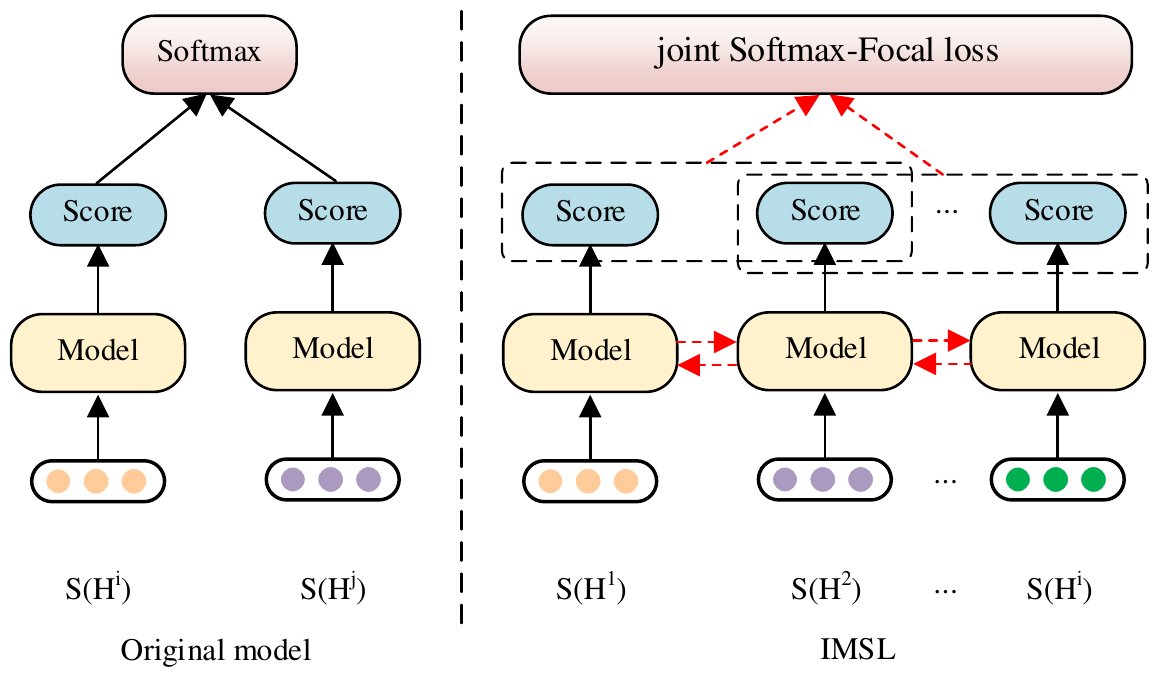}
\end{center}
\caption{Comparison of the interaction between the traditional model and the IMSL model, where S ($\rm H^i$) represents the input sequence containing the i-th hypothesis} \label{fig:comparison}
\end{figure}

Based on the above reasoning, we propose a new abductive reasoning model  called Interactive Model with Structural Loss (IMSL) for $\alpha$NLI as shown in Figure \ref{fig:comparison}.
The IMSL model mainly consists of two components: interactive model and structural loss. On the one hand, note that in the process of extracting the language features of an arbitrary hypothesis, its relationship to other hypotheses should also be considered because the  hypotheses are often semantically related \citep{pearl1986evidential}. To this end, we can design an information interaction layer to capture the relationship between different hypotheses and produce more discriminative language feature vectors. On the other hand, we have constructed a new loss function called ``joint softmax focal loss'' inspired by a recent work  \citep{DBLP:conf/iccv/LinGGHD17}. It is essentially a structural softmax based Focal loss formed by sequentially constructing a loss for each score group that composed by a correct hypothesis and all wrong hypotheses. When compared with conventional models, we argue that IMSL is more powerful for the task of $\alpha$NLI by jointly exploiting the rich relation among competing hypotheses. The main technical contributions of this work can be summarized as follows.

1) For $\alpha$NLI task, we claim that, the correct hypotheses of a given observation pair are often diverse, and there is no need to tell them apart. The wrong hypotheses contribute the same to the task. We regroup instead of ranking all hypotheses, as shown in Figure \ref{fig:method}.

2) Aiming at the problem of incorrect probability distribution between correct hypotheses in the training process, a joint softmax focal loss is proposed. For the hypotheses groups formed in the rearrange process, we design a softmax-based focal loss for each group and combine them into a joint loss.

3) In view of the problem that traditional models cannot capture the language relationship between different hypotheses, we have added an information interaction layer between different hypothesis models. The information interaction layer increases the area under the receiver's operating characteristic curve (AUC) by about 5\%.

4) Impressive abductive reasoning performance is achieved by IMSL when tested using RoBERTa as the pretrained language model. The best language model DeBERTa \citep{DBLP:conf/iclr/HeLGC21} is not tested due to the constraint by our limited GPU resources (4-piece RXT 2080Ti). In our experiment, compared with all recent algorithms whose codes have been made publicly available, the IMSL method has achieved state-of-the-art results in ACC and AUC on both the validation set and test set. Besides, on the public leaderboard\footnote{\url{https://leaderboard.allenai.org/anli/submissions/public}}, IMSL is the best non-DeBERTa based algorithm and ranks 4/56 in all (including both DeBERTa based and non-DeBERTa based) competing methods. 

\section{Related Work}

$\alpha$NLI task solves an abductive reasoning problem based on natural language inference (NLI). In the past years, there has been an explosion of NLI benchmarks, since the Recognizing Textual Entailment (RTE) Challenges was introduced by \cite{DBLP:conf/mlcw/DaganGM05} in the early 2000s. Then, in order to find the most reasonable explanation of the incomplete observations, \cite{DBLP:conf/iclr/BhagavatulaBMSH20} studied the feasibility of language-based abductive reasoning and proposed the task of $\alpha$NLI. It pays more attention to the information provided in the premise than the RTE task. For traditional RTE, the main task is to judge the relationship between the premise sentence and the hypothetical sentence, but the main objective of $\alpha$NLI is to select the most plausible hypothesis among the hypotheses given two observations.

$\alpha$NLI is the first language-based abductive reasoning study. This shift from logic-based to language-based reasoning draws inspirations from a significant body of works on language-based entailment \citep{DBLP:conf/emnlp/BowmanAPM15}; \citep{DBLP:conf/naacl/WilliamsNB18}, language-based logic \citep{lakoff1970linguistics}; \citep{DBLP:conf/acl/MacCartneyM07}, and language-based commonsense reasoning \citep{DBLP:conf/naacl/MostafazadehCHP16}; \citep{DBLP:conf/emnlp/ZellersBSC18}. In addition to establish $\alpha$NLI, \cite{DBLP:conf/mlcw/DaganGM05} have also released a new challenge dataset, i.e., ART, which can be visited through the first footnote in this paper. The authors have also formulate the task as a multiple-choice task to support easy and reliable automatic evaluation. Specifically, from a given context, the task is to choose the more reliable explanation from a given pair of hypotheses choices. 

However, discriminating correct from wrong does not measure the plausibility of a hypothesis in $\alpha$NLI \citep{DBLP:conf/sigir/ZhuPLC20}. So, to fully model the plausibility of the hypotheses, Zhu et al. turn to the perspective of ranking and propose a novel learning to rank for reasoning ($\rm L2R^2$) approach for the task. The authors rank the hypotheses based on the number of times they appear in the dataset, and use some pairwise rankings as well as a listwise ranking as loss. Pairwise rankings contains Ranking SVM \citep{herbrich2000large}, RankNet\citep{DBLP:conf/icml/BurgesSRLDHH05}, LambdaRank\citep{DBLP:conf/nips/BurgesRL06}, and Listwise Ranking contains ListNet\citep{DBLP:conf/icml/CaoQLTL07}, ListMLE
\citep{DBLP:conf/icpr/LiL0R20} and ApproxNDCG\citep{DBLP:journals/ir/QinLL10}. The experiments on the ART dataset show that reformulating the $\alpha$NLI task as ranking task really brings obvious improvements. After that, \cite{DBLP:conf/emnlp/PaulF20} proposes a novel multi-head knowledge attention model that encodes semi-structured commonsense inference rules and learns to incorporate them in a transformer based reasoning cell. The authors still prove that a model using counterfactual reasoning is useful for predicting abductive reasoning tasks. Accordingly, they have established a new task called Counterfactual Invariance Prediction (CIP) and provide a new dataset for this. 

In addition to the abductive reasoning models, the pre-trained language model still plays an important role in $\alpha$NLI task. Early ways for language inference are constructed directly by some simple statistical measures like bag-of-words and word matching. Later, various kinds of neural network architectures are used to discover useful features in the languages, like word2vec\citep{DBLP:journals/corr/abs-1301-3781}  and GloVe\citep{DBLP:conf/emnlp/PenningtonSM14}. Recent works have developed contextual word representation models, e.g.,Embeddings from Language Models (ELMO) by \cite{DBLP:conf/naacl/PetersNIGCLZ18} and Bidirectional Encoder Representations from Transformers(BERT) by \cite{DBLP:conf/naacl/DevlinCLT19}. The original implementation and architecture of BERT has been outperformed by several variants and other transformer-based models, such as RoBERTa, DeBERTa and UNIMO. RoBERTa\citep{DBLP:journals/corr/abs-1907-11692} replaces training method of BERT and uses larger batches and more data for training. DeBERTa\citep{DBLP:conf/iclr/HeLGC21} uses the disentangled attention mechanism and an enhanced mask decoder to improves the BERT and RoBERTa models. In order to effectively adapt to unimodal and multimodal understanding task, \cite{DBLP:conf/acl/LiGNXLL0020} proposes the UNIMO model. In this paper, however, restricted by our computing resources, RoBERTa is selected as our language model.

\section{Interactive Model with Structural Loss (IMSL) Method}

IMSL model consists of two components: context coding layer and information interaction layer (the backbone network) as well as a joint softmax focal loss (objective function). The design of the model architecture and loss function are described in detail below.

\subsection{information interaction Model}

\begin{figure}[htb]
\begin{center}
\includegraphics[width=120mm]{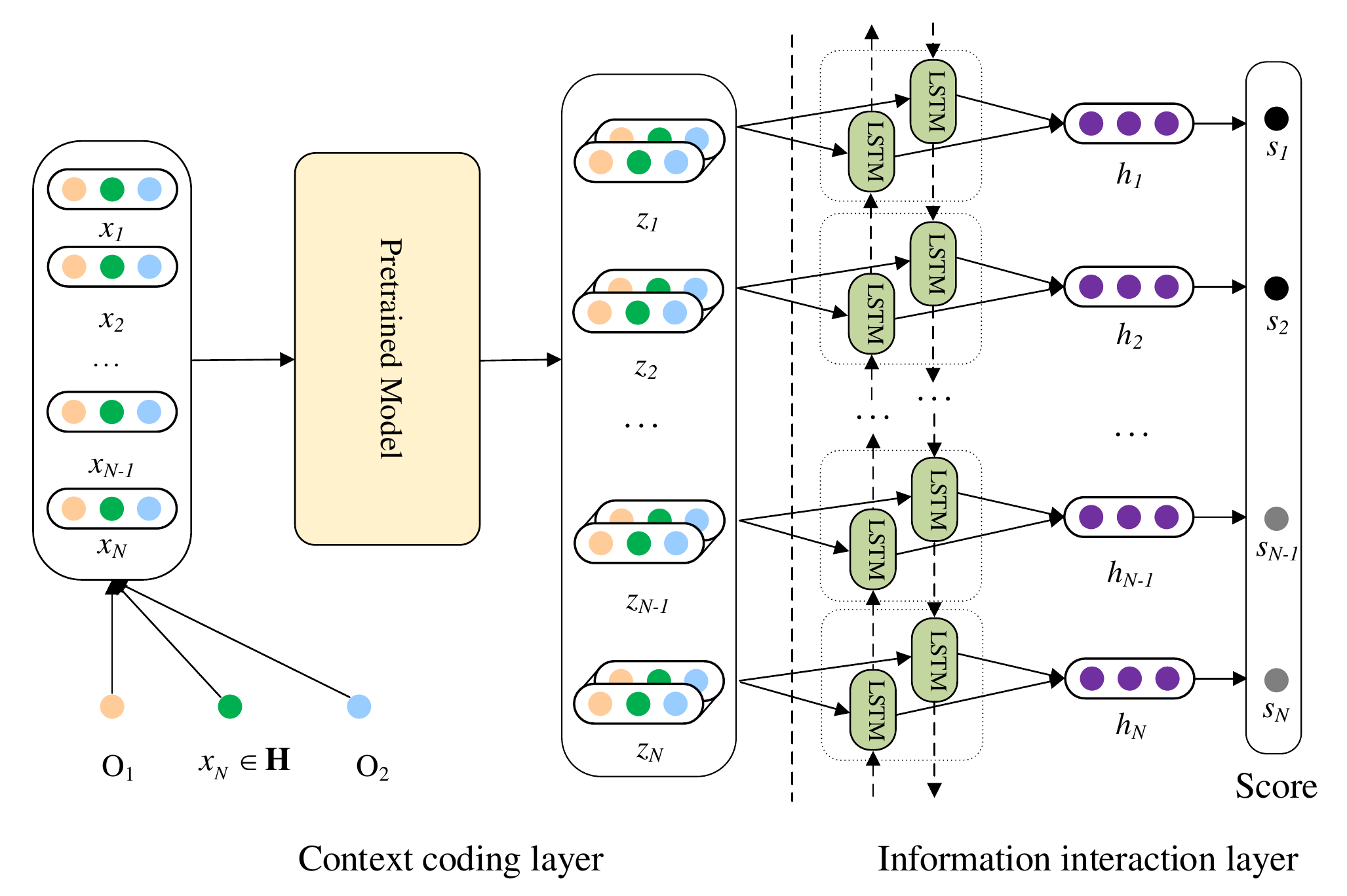}
\end{center}
\vspace{-0.1in}
\caption{The proposed IMSL model consists of a context coding layer (using a pre-trained model) and an information interaction layer (characterizing the relationship among different hypotheses).} \label{fig:model}
\vspace{-0.1in}
\end{figure}

\textbf{Model input:} Under the framework of IMSL, a training sample $X$ includes two given observations (i.e., $\rm O_1$ and $\rm O_2$) and a group of  candidate hypotheses denoted by $\mathbf{H}=\left\{\rm{H^{j}}\right\}_{j=1}^{N}$ ($N$ is the number of candidate hypotheses). Then, binary labels $\textbf{y}=\left\{y_{j}\right\}_{j=1}^{N}$ are assigned to each hypothesis by $y_{j}=1$ when $\rm{H}^j$ is the correct, and $y_{j}=0$ when $\rm{H}^j$ is the wrong. The task of abductive reasoning can be characterized by a mapping from $X$ to $\textbf{y}$.
For explicitly estimating the relation between each hypothesis and the two observations, we can construct a  triad for each hypothesis as $x_{j}=\left[\mathrm{O}_{1} ; \rm{H^j} ; \mathrm{O}_{2}\right]\left(\rm{H^j} \in \mathbf{H}\right)$. This way, each sample in the training set $X$ can be represented by $[x_{1},x_{2},\cdots,x_{N}]\rightarrow [y_{1},y_{2},\cdots,y_{N}]$.

\textbf{Context coding layer:} We use a pre-trained language model (RoBERTa-large is used in our experiment) to calculate the contextual representation of the text. For each word in a single input $x_j$, an embedding vector with context information is generated. For each sentence in a single input  $x_j$, a sentence-level embedding matrix $v_{j}=\operatorname{encode}\left(x_{j}\right)$ is first obtained, where $\operatorname{encode(\cdot)}$ denotes the pre-trained model for encoding. Then we can sum the word embedding dimensions in the feature matrix to generate the feature vector $z_j$.

\textbf{Information interaction layer:} Traditional models only consider one single input $x_j$ during scoring as shown in Fig. \ref{fig:comparison}, which makes it difficult to capture the relationship between different inputs (e.g., $x_j$ and $x_{j+1}$). To exploit the dependency between two different inputs, we propose to construct a novel information interaction layer as follows. First, a pair of feature vectors $z_j$ and $z_{j+1}$ can be generated after $x_j$ and $x_{j+1}$ are passed through the context encoding layer. Second, we plug  $z_j$ and $z_{j+1}$ into the information interaction layer and use BiLSTM to acquire the distributed feature representation $f_j$ and $f_{j+1}$. Finally, a fully connected module outputs the corresponding scores $s_j$ and $s_{j+1}$. A flowchart of the context coding and information interaction layers is shown in Figure \ref{fig:model}.

To efficiently use contextual information, we use $z_j$ as the input of BiLSTM, which aim at exploiting the dependency relationship between the feature vectors. BiLSTM uses a forward LSTM and a backward LSTM for each sequence to obtain two separate hidden states: $\overrightarrow{{h}_{j}}, \overleftarrow{h}_{j}$. The key idea of BiLSTM is to generate the final output at time $t$ by concatenating these two hidden states:
\vspace{-0.02in}
\begin{equation}
\begin{aligned}
{h}_{j}=\left[\overrightarrow{{h}_{j}}, \overleftarrow{{h}}_{j}\right].
\end{aligned}
\end{equation}
\vspace{-0.02in}
After passing the BiLSTM layer, we can get a bidirectional hidden state vector $h_j$, then use the fully connected layer to generate the final score $s_j$. For computational efficiency, a linear regression formula is adopted here for prediction score:
\begin{equation}
\begin{aligned}
s_{j}=W_{j} \cdot h_{j}+b_{j},
\end{aligned}
\end{equation}
where $W_{j} \in \mathbb{R}^{2 d \times d}, b_{j} \in \mathbb{R}^{d}$.

\subsection{Joint Softmax focal Loss function}

\begin{figure}[htb]
\begin{center}
\includegraphics[width=100mm]{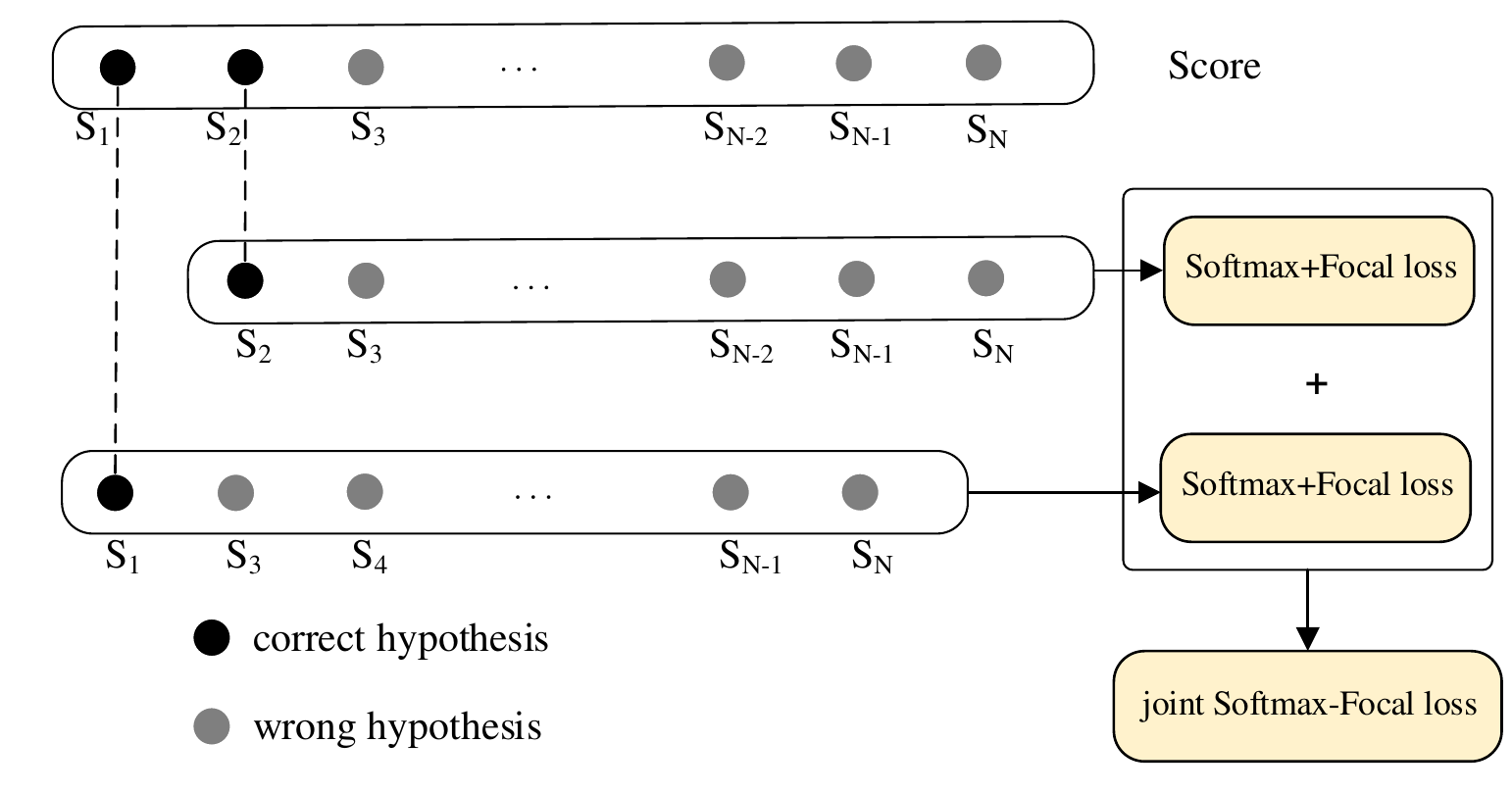}
\end{center}
\caption{An example of rearrange groups for joint softmax focal loss.} \label{fig:cross}
\end{figure}

Based on the output score layer of the IMSL model, we propose to design a new structural loss function based on the principle of abductive reasoning.  Instead of ranking-based approach, the proposed loss function for each sample is formulated as a linear combination of softmax focal losses for several rearranged groups, which is called joint softmax focal loss. The detailed procedure of generating multiple rearranged groups is shown in Figure \ref{fig:cross}. We note that it is unnecessary to compare the group of correct hypotheses; while the exploration of the relationship between correct hypothesis and wrong hypotheses is sufficient for the task of score prediction. Therefore, we can rearrange the set of $N$ prediction scores into several groups, each of which only contains a single correct hypothesis. A toy example is given in Figure \ref{fig:cross} where the two hypotheses are correct, and all other hypotheses are wrong. In this example, the total $N$ scores can be divided into two groups associated with two correct hypotheses, respectively. 

With the above construction, we can first apply the softmax operation to each rearranged group and then combine them into a joint loss function. In the first step, each prediction score is given by:
\begin{equation}
\begin{aligned}
{\hat y_n} = \left\{ \begin{array}{l}
\frac{{{e^{s_n^1}}}}{{{e^{s_n^1}} + \sum\nolimits_i {{e^{s_i^0}}} }},\qquad \quad \ \ if\quad{y_n} = 1.\\
\sum\nolimits_j {\frac{{{e^{s_n^0}}}}{{K\left( {{e^{s_j^1}} + \sum\nolimits_i {{e^{s_i^0}}} } \right)}}} ,if\quad{y_n} = 0.
\end{array} \right.
\end{aligned}
\end{equation}
where $y_n$ is the correct/wrong label, and $\hat{y}_{n}$ represents a predicted value. The normalization factor $K = \sum\nolimits_j {{y_j}}$ represents the number of correct hypotheses. Note that $s_{i}^{0}$ represents the scores of the wrong hypotheses, where $i$ is the position of the false label. Similarly, $s_{i}^{1}$ indicates the score of the correct hypotheses.

In addition to the softmax loss, we have borrowed the idea of focal loss from \cite{DBLP:conf/iccv/LinGGHD17} and introduce a balancing factor $a \in (0,1)$ to control the shared weight of the correct hypothesis and the wrong ones. Here, $a$ is used for the correct hypotheses, and $1-a$ is used for the wrong hypotheses, i.e.,
\begin{equation}
\begin{aligned}
\beta_{n}=y_{n} \cdot a+\left(1-y_{n}\right)(1-a).
\end{aligned}
\label{eq:bal}
\end{equation}
Putting things together, we can rewrite the joint softmax focal loss as 
\begin{equation}
\begin{aligned}
{\cal L }=F_{l}(y, \hat{y})=-\sum_{n} \beta_{n} \cdot\left(1-p_{n}\right)^{\gamma} \cdot \log \left(p_{n}\right).
\end{aligned}
\label{eq:focal}
\end{equation}
where 
\begin{equation}
\begin{aligned}
p_{n}=y_{n} \cdot \hat{y}_{n}+\left(1-y_{n}\right)\left(1-\hat{y}_{n}\right)+\varepsilon.
\end{aligned}
\end{equation}
Here, the parameter $\gamma$ is included for regulating the model's attention to hard hypotheses during the training of IMSL model. As suggested in \cite{DBLP:conf/iccv/LinGGHD17}, $\gamma \in [0.5,5]$. Then, a small positive real number $\varepsilon$ of 1\emph{e}-8 is used to avoid the numerical instability problem. In practice, both $a$ and $\gamma$ are used as hyper-parameters which can be tuned by cross-validation. 
For the example shown in Figure \ref{fig:cross}, by assuming that the softmax focal losses for the two groups are ${\cal L}_{group1} $ and ${\cal L}_{group2} $, we can obtain the overall loss by ${\cal L}_{sample} = {\cal L}_{group1} + {\cal L}_{group2} $. Furthermore, the total loss for all  training samples can be estimated by the sum of the losses over individual samples.

\section{Experiment}
In this section, the experimental results on public data sets are presented to evaluate the method proposed in this paper.

In recent years, more and more Pre-Training models have been proposed, such as DeBERTa \citep{DBLP:conf/iclr/HeLGC21}, UNIMO \citep{DBLP:conf/acl/LiGNXLL0020}, etc. They use more data for training and have more parameters. Due to limited computational resources, we did not conduct comparative experiments with these high-performing yet computationally demanding pretrained models.

\textbf{Evaluation indicators:} AUC and ACC are the most common evaluation indicators. Since the original ACC cannot evaluate the model that is far away from the test data, AUC is added as an additional evaluation index to handle skewed sample distribution. AUC is a measurement method that is statistically consistent and more discriminative than ACC.

\subsection{Experimental setup}
Tasks and settings: The $\alpha$NLI task uses the ART dataset, which is the first large-scale benchmark dataset used for abductive reasoning in narrative texts. It consists of about 20,000 observations and about 200,000 pairs of hypotheses. The observations come from a collection of manually curated stories, and the hypotheses are collected through crowdsourcing. In addition, the candidate hypotheses for each narrative context in the test set are selected through the adversarial filtering algorithm with BERT-L (Large) as the opponent. The input and output formats are shown in Table \ref{table:form}.
\begin{table}[htb]
\caption{The format of data input and output in $\alpha$NLI task} \label{table:form}
\begin{center}
\begin{tabular}{c|c|c}
\hline
Task & Input Format                                     & Output  Format \\ \hline
$\alpha$NLI & {[}CLS{]} $\rm O_1$ {[}SEP{]} $\rm H^i$ {[}SEP{]} $\rm O_2$ {[}SEP{]} & $\rm H^1$ or $\rm H^2$ \\ \hline
\end{tabular}
\end{center}
\end{table}

\textbf{Hyperparameter details:} Due to the difference in the amount of data, the focusing parameter and the amount of training data will vary. For different training data, select the hyperparameter that produces the best performance on the test set. Specifically, the learning rate is fixed at 1e-6, the batch size is fixed at 1, and the training batch will vary with the amount of training data. Training uses Cross Softmax+Focal Loss. For the validation set, ACC and AUC are used for evaluation. Use the results of five different seeds to evaluate the performance of the test set.

\textbf{Baseline:} We have used the following four baseline models for comparison:
A) BERT \citep{DBLP:conf/naacl/DevlinCLT19} is a language model that uses a masked language model and predicts the next sentence as the target training. For example, it masks certain words in the input, and then trains it and predicts the words that are blocked.
B) RoBERTa \citep{DBLP:journals/corr/abs-1907-11692} has the same structure as BERT, but there is no prediction (NSP) for the next sentence. RoBERTa-B (ase) and RoBERTa-L (arge) use more data and larger batches for training.
C) Learning to Rank for Reasoning ($\rm L2R^2$) \citep{DBLP:conf/sigir/ZhuPLC20} reprogrammed the $\alpha$NLI task as a ranking problem, using a learning ranking framework that includes a score function and a loss function.
D) Multi-Head Knowledge Attention (MHKA) \citep{DBLP:conf/emnlp/PaulF20} proposed a new multihead knowledge attention model, and used a novel knowledge integration technology.

\subsection{Experimental results}
Our experimental results in the $\alpha$NLI task are shown in Table \ref{table:results}. The baseline comparison models are: Majority, GPT, BERT-L, RoBERTa-L, $\rm L2R^2$ and MHKA related results. It can be observed that the IMSL method improves about 3.5\% in ACC and about 7.5\% in AUC compared with RoBERTa-L. The results show that the improvement of ACC is mainly attributed to the new IMSL loss function, and the improvement of AUC is mainly attributed to the exploitation of the relationship between the hypotheses by the proposed information interaction layer.
\begin{table}[htb]
\caption{Results on the $\alpha$NLI task: The results are quoted from \cite{DBLP:conf/iclr/BhagavatulaBMSH20}, L=Large} \label{table:results}
\begin{center}
\begin{tabular}{llll}
\hline
Model                 & Dev(ACC\%)     & Dev(AUC\%)     & Test(ACC\%) \\ \hline
Human Perf            & -              & \textbf{-}     & 91.40       \\ \hline
Majority              & 50.80          & -              & -           \\
GPT                   & 62.70          & -              & 62.30       \\
BERT-L                & 69.10          & 69.03          & 68.90       \\
RoBERTa-L             & 85.76          & 85.02          & 84.48       \\
$\rm L2R^2$           & 88.44          & 87.53          & 86.81       \\
MHKA                  & 87.85          & -              & 87.34       \\
\rowcolor[HTML]{E7E6E6} 
Ours                  &                &                &             \\
RoBERTa-L+IMSL & \textbf{89.20}          & \textbf{92.50} & \textbf{87.83}       \\ \hline

\end{tabular}
\end{center}
\end{table}

\begin{figure}[htb]
\begin{center}
\includegraphics[width=105mm]{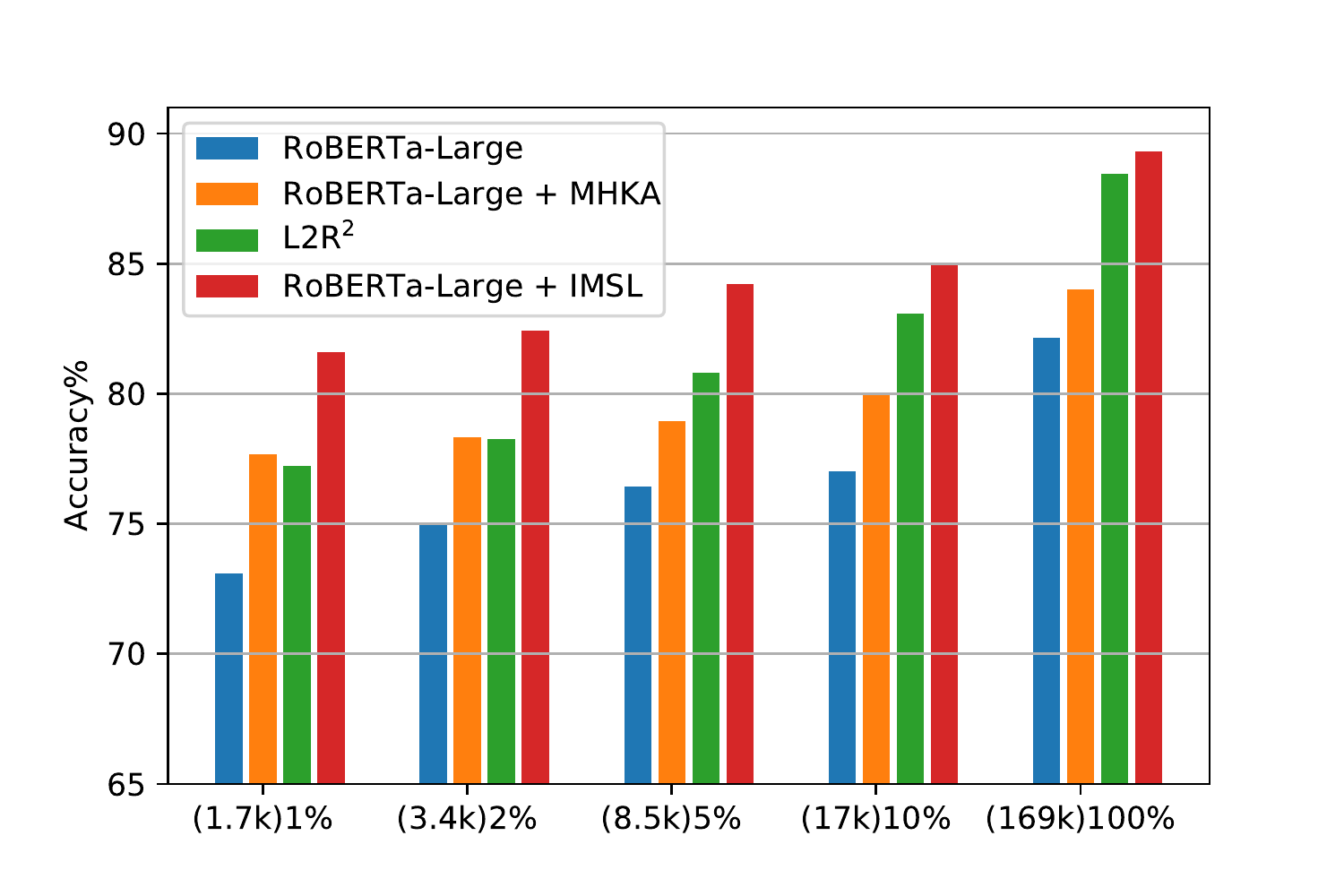}
\end{center}
\vspace{-0.1in}
\caption{The accuracy of $\alpha$NLI under low resource settings}
\vspace{-0.1in}
\label{fig:lowset}
\end{figure}

\textbf{Low-resource setting}: Testing the robustness of the model to sparse data on $\alpha$NLI tasks refers to the low-resource scenario where the MHKA model uses \{1,2,5,10,100\}\% training data respectively. Figure \ref{fig:lowset} shows how the model improves on MHKA, RoBERTa-Large, and $\rm L2R^2$. The experimental results show that the model in this paper can achieve better results in the case of low resource setting. When using 1\% training data only, the improvement brought by IMSL is the most significant, which is about 4\% higher than that of $\rm L2R^2$ and MHKA. Experimental results show that our method performs consistently better than other competing methods on low-resource data sets.
\section{Detailed analysis via ablation study}
To more clearly show the contribution of each module, we have done corresponding comparative experiments on both information interaction layer and hyperparameter tuning.
\subsection{Ablation study of the information interaction layer}

First, we have conducted ablation study experiments on the related BiLSTM to investigate the role played by the information interaction layer. The hyperparameters of Focal Loss will be fixed to reduce the impact on BiLSTM. Through the experimental results, it can be found that the addition of BiLSTM greatly improves the AUC, but does not have a significant impact on ACC. The following Figure \ref{fig:nointer} shows the visualization results on the validation set. In the plot, the abscissa is the score of hypothesis 1 and the ordinate is the score of hypothesis 2. The red points in the upper left corner correspond to the subset of correct hypotheses, so do the blue points in the lower right corner. It can be seen that the introduction of the information interaction layer pushes all points further away from each other and toward the four corners. It follows that the margin between the positive and negative samples is larger, implying improved discriminative power.

\begin{figure}[htb]
\begin{center}
\subfigure[using the information interaction layer ]{\includegraphics[width=55mm]{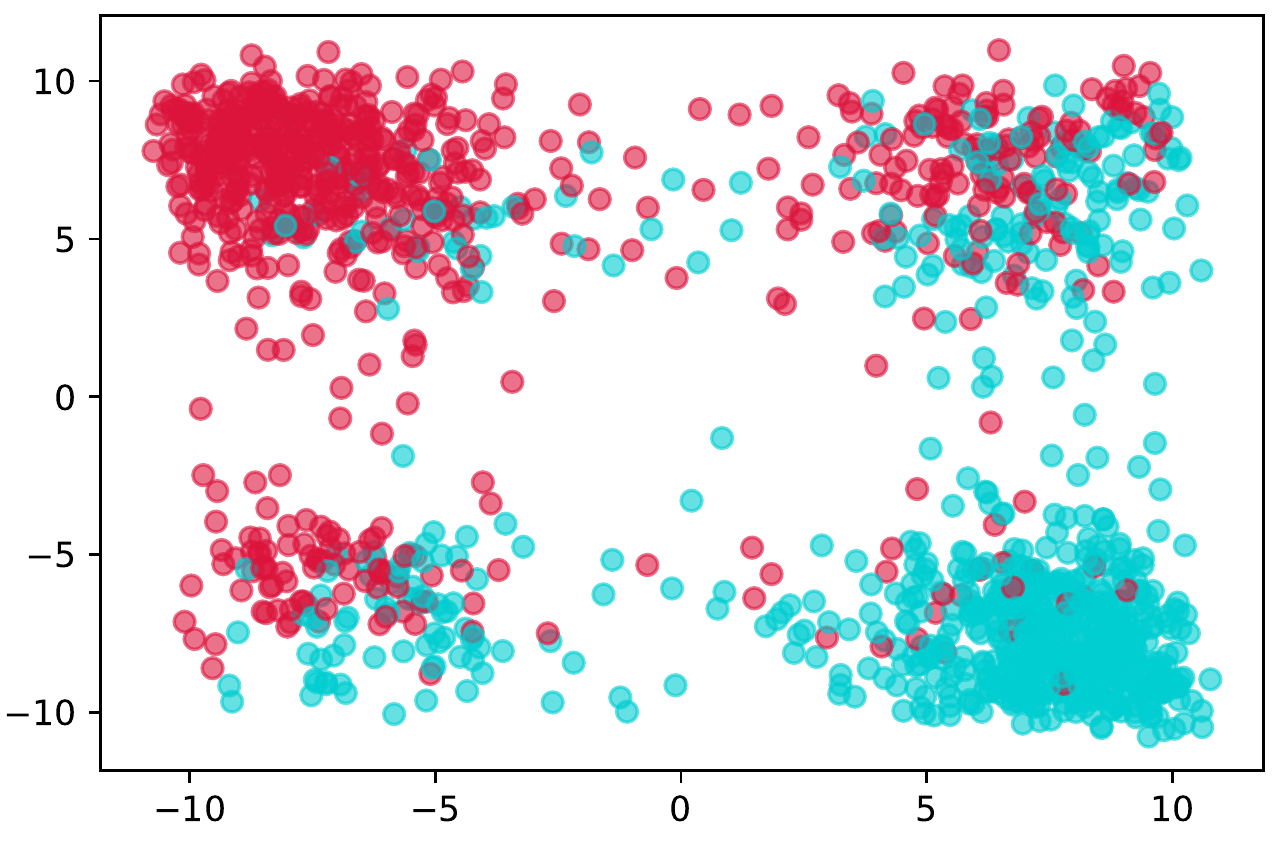}}
\quad
\subfigure[without the information interaction layer ]{\includegraphics[width=55mm]{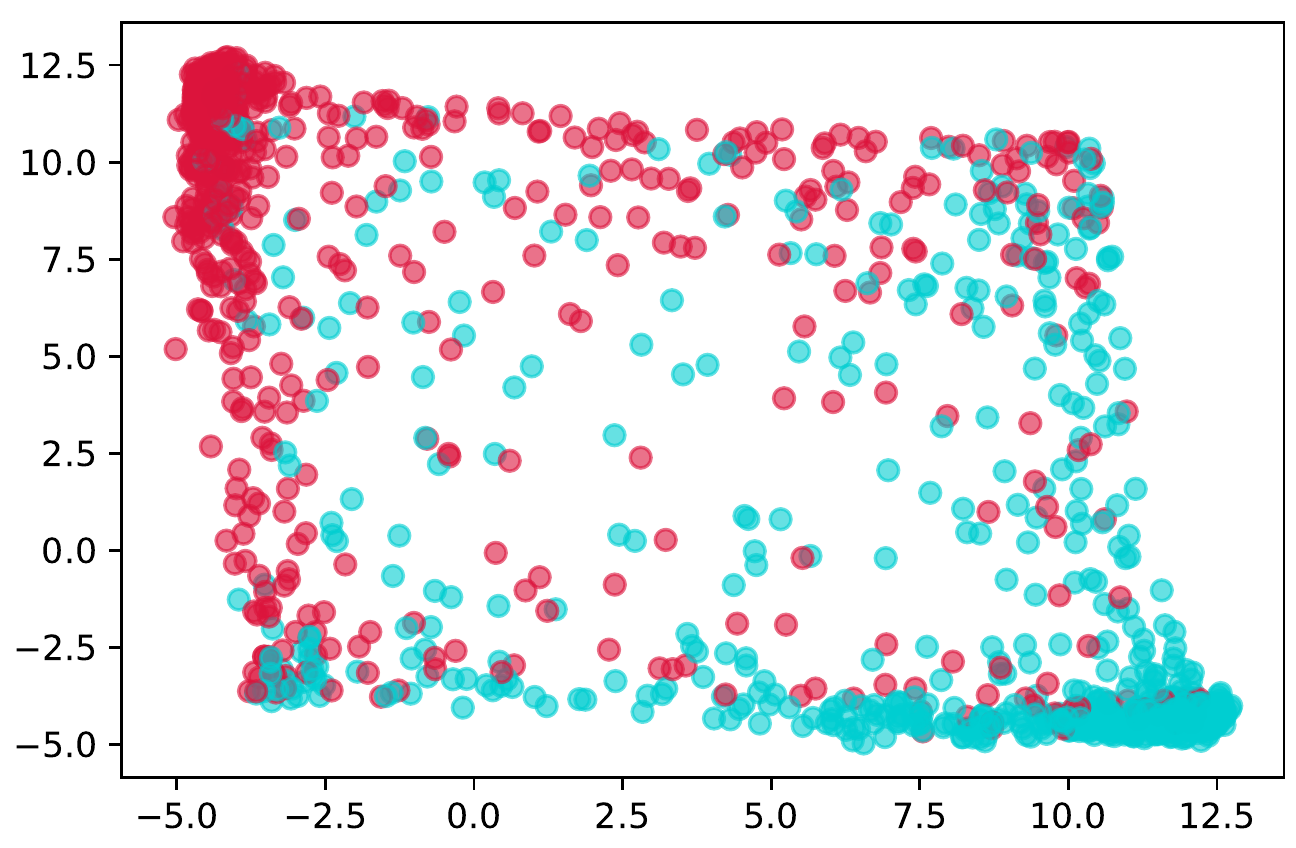} }

\end{center}
\caption{Score distribution using and without using the information interaction layer} \label{fig:nointer}
\end{figure}

\subsection{Parameter comparison}
When we select the parameters of Focal Loss, several experiments were carried out on the two hyperparameters of the balancing factor and the focusing parameter.
The focusing parameter $\gamma$ in Eq. \eqref{eq:focal} can automatically down-weight the contribution of easy examples during training and rapidly focus the model on hard examples; while the balancing factor $0<a<1$ in Eq. \eqref{eq:bal} controls the tradeoff between correct and incorrect hypotheses. Figure \ref{fig:heat} below shows the ACC performance of IMSL model with different focusing parameters and balance factors. In this study, \{1, 2, 3\} is used as the option of focusing parameter, and \{0.45, 0.5, 0.55\} is used as the set of balancing factor. It can be observed that as the most effective parameter couple is given by $\gamma= 2$, $a = 0.55$.
\begin{figure}[htb]
\begin{center}
\includegraphics[width=75mm]{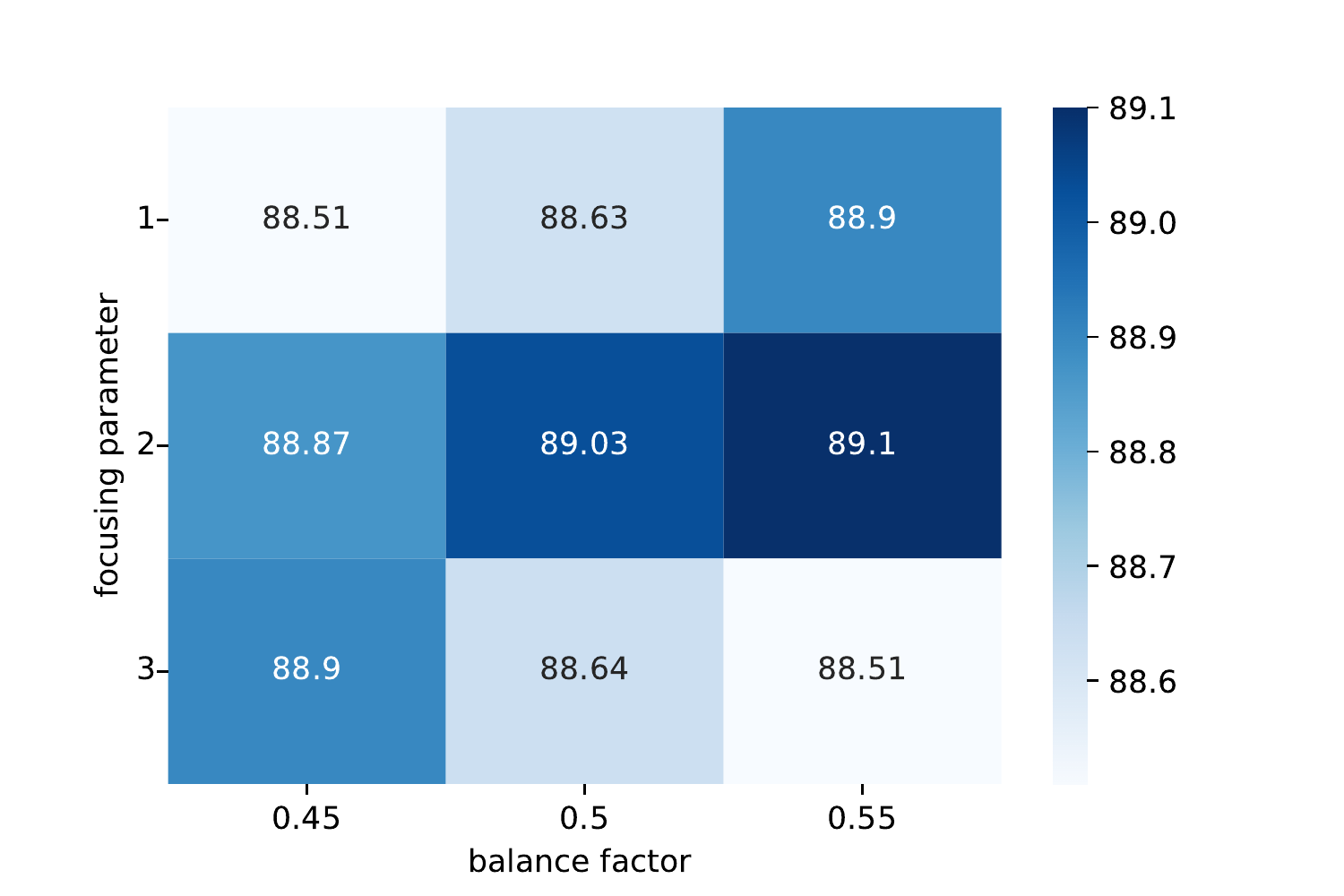}
\end{center}
\caption{The ACC results of adjusting the balance factor $\gamma$ and focusing parameter $a$.} \label{fig:heat}
\end{figure}
\section{Summary}

In this paper, an IMSL method is proposed for commonsense abductive reasoning. It includes an information interaction layer that captures the relationship between different hypotheses, and a joint loss for our proposed way of grouping the correct/wrong hypotheses. Experimental results show that on $\alpha$NLI tasks, IMSL has better performance on ACC and AUC, especially in low-resource settings, IMSL can significantly improve the accuracy. 
\bibliography{main}
\bibliographystyle{main}
\end{document}

%% file: math_commands.tex
\usepackage{amsmath,amsfonts,bm}

\def\eqref#1{equation~\ref{#1}}

\def\1{\bm{1}}

\DeclareMathAlphabet{\mathsfit}{\encodingdefault}{\sfdefault}{m}{sl}
\SetMathAlphabet{\mathsfit}{bold}{\encodingdefault}{\sfdefault}{bx}{n}













%% file: main.bbl
\begin{thebibliography}{32}
\providecommand{\natexlab}[1]{#1}
\providecommand{\url}[1]{\texttt{#1}}
\expandafter\ifx\csname urlstyle\endcsname\relax
  \providecommand{\doi}[1]{doi: #1}\else
  \providecommand{\doi}{doi: \begingroup \urlstyle{rm}\Url}\fi

\bibitem[Bauer \& Bansal(2021)Bauer and Bansal]{DBLP:conf/eacl/BauerB21}
Lisa Bauer and Mohit Bansal.
\newblock Identify, align, and integrate: Matching knowledge graphs to
  commonsense reasoning tasks.
\newblock In \emph{Proceedings of the 16th Conference of the European Chapter
  of the Association for Computational Linguistics: Main Volume, {EACL} 2021,
  Online, April 19 - 23, 2021}, pp.\  2259--2272. Association for Computational
  Linguistics, 2021.

\bibitem[Bhagavatula et~al.(2020)Bhagavatula, Bras, Malaviya, Sakaguchi,
  Holtzman, Rashkin, Downey, Yih, and Choi]{DBLP:conf/iclr/BhagavatulaBMSH20}
Chandra Bhagavatula, Ronan~Le Bras, Chaitanya Malaviya, Keisuke Sakaguchi, Ari
  Holtzman, Hannah Rashkin, Doug Downey, Wen{-}tau Yih, and Yejin Choi.
\newblock Abductive commonsense reasoning.
\newblock In \emph{8th International Conference on Learning Representations,
  {ICLR} 2020, Addis Ababa, Ethiopia, April 26-30, 2020}. OpenReview.net, 2020.

\bibitem[Bowman et~al.(2015)Bowman, Angeli, Potts, and
  Manning]{DBLP:conf/emnlp/BowmanAPM15}
Samuel~R. Bowman, Gabor Angeli, Christopher Potts, and Christopher~D. Manning.
\newblock A large annotated corpus for learning natural language inference.
\newblock In \emph{Proceedings of the 2015 Conference on Empirical Methods in
  Natural Language Processing, {EMNLP} 2015, Lisbon, Portugal, September 17-21,
  2015}, pp.\  632--642. The Association for Computational Linguistics, 2015.

\bibitem[Burges et~al.(2005)Burges, Shaked, Renshaw, Lazier, Deeds, Hamilton,
  and Hullender]{DBLP:conf/icml/BurgesSRLDHH05}
Christopher J.~C. Burges, Tal Shaked, Erin Renshaw, Ari Lazier, Matt Deeds,
  Nicole Hamilton, and Gregory~N. Hullender.
\newblock Learning to rank using gradient descent.
\newblock In \emph{Machine Learning, Proceedings of the Twenty-Second
  International Conference {(ICML} 2005), Bonn, Germany, August 7-11, 2005},
  volume 119 of \emph{{ACM} International Conference Proceeding Series}, pp.\
  89--96. {ACM}, 2005.

\bibitem[Burges et~al.(2006)Burges, Ragno, and Le]{DBLP:conf/nips/BurgesRL06}
Christopher J.~C. Burges, Robert Ragno, and Quoc~Viet Le.
\newblock Learning to rank with nonsmooth cost functions.
\newblock In \emph{Advances in Neural Information Processing Systems 19,
  Proceedings of the Twentieth Annual Conference on Neural Information
  Processing Systems, Vancouver, British Columbia, Canada, December 4-7, 2006},
  pp.\  193--200. {MIT} Press, 2006.

\bibitem[Cao et~al.(2007)Cao, Qin, Liu, Tsai, and Li]{DBLP:conf/icml/CaoQLTL07}
Zhe Cao, Tao Qin, Tie{-}Yan Liu, Ming{-}Feng Tsai, and Hang Li.
\newblock Learning to rank: from pairwise approach to listwise approach.
\newblock In \emph{Machine Learning, Proceedings of the Twenty-Fourth
  International Conference {(ICML} 2007), Corvallis, Oregon, USA, June 20-24,
  2007}, volume 227 of \emph{{ACM} International Conference Proceeding Series},
  pp.\  129--136. {ACM}, 2007.

\bibitem[Chen et~al.(2017)Chen, Zhu, Ling, Wei, Jiang, and
  Inkpen]{DBLP:conf/acl/ChenZLWJI17}
Qian Chen, Xiaodan Zhu, Zhen{-}Hua Ling, Si~Wei, Hui Jiang, and Diana Inkpen.
\newblock Enhanced {LSTM} for natural language inference.
\newblock In \emph{Proceedings of the 55th Annual Meeting of the Association
  for Computational Linguistics, {ACL} 2017, Vancouver, Canada, July 30 -
  August 4, Volume 1: Long Papers}, pp.\  1657--1668. Association for
  Computational Linguistics, 2017.

\bibitem[Dagan et~al.(2005)Dagan, Glickman, and
  Magnini]{DBLP:conf/mlcw/DaganGM05}
Ido Dagan, Oren Glickman, and Bernardo Magnini.
\newblock The {PASCAL} recognising textual entailment challenge.
\newblock In \emph{Machine Learning Challenges, Evaluating Predictive
  Uncertainty, Visual Object Classification and Recognizing Textual Entailment,
  First {PASCAL} Machine Learning Challenges Workshop, {MLCW} 2005,
  Southampton, UK, April 11-13, 2005, Revised Selected Papers}, volume 3944 of
  \emph{Lecture Notes in Computer Science}, pp.\  177--190. Springer, 2005.

\bibitem[Devlin et~al.(2019)Devlin, Chang, Lee, and
  Toutanova]{DBLP:conf/naacl/DevlinCLT19}
Jacob Devlin, Ming{-}Wei Chang, Kenton Lee, and Kristina Toutanova.
\newblock {BERT:} pre-training of deep bidirectional transformers for language
  understanding.
\newblock In \emph{Proceedings of the 2019 Conference of the North American
  Chapter of the Association for Computational Linguistics: Human Language
  Technologies, {NAACL-HLT} 2019, Minneapolis, MN, USA, June 2-7, 2019, Volume
  1 (Long and Short Papers)}, pp.\  4171--4186. Association for Computational
  Linguistics, 2019.

\bibitem[He et~al.(2021)He, Liu, Gao, and Chen]{DBLP:conf/iclr/HeLGC21}
Pengcheng He, Xiaodong Liu, Jianfeng Gao, and Weizhu Chen.
\newblock Deberta: decoding-enhanced bert with disentangled attention.
\newblock In \emph{9th International Conference on Learning Representations,
  {ICLR} 2021, Virtual Event, Austria, May 3-7, 2021}. OpenReview.net, 2021.

\bibitem[Herbrich et~al.(2000)Herbrich, Graepel, Obermayer,
  et~al.]{herbrich2000large}
Ralf Herbrich, Thore Graepel, Klaus Obermayer, et~al.
\newblock Large margin rank boundaries for ordinal regression.
\newblock \emph{Advances in large margin classifiers}, 88\penalty0
  (2):\penalty0 115--132, 2000.

\bibitem[Huang et~al.(2020)Huang, Zhang, Elachqar, and
  Cheng]{DBLP:conf/acl/HuangZEC20}
Yichen Huang, Yizhe Zhang, Oussama Elachqar, and Yu~Cheng.
\newblock {INSET:} sentence infilling with inter-sentential transformer.
\newblock In \emph{Proceedings of the 58th Annual Meeting of the Association
  for Computational Linguistics, {ACL} 2020, Online, July 5-10, 2020}, pp.\
  2502--2515. Association for Computational Linguistics, 2020.

\bibitem[Lakoff(1970)]{lakoff1970linguistics}
George Lakoff.
\newblock Linguistics and natural logic.
\newblock \emph{Synthese}, 22\penalty0 (1):\penalty0 151--271, 1970.

\bibitem[Li et~al.(2020)Li, Liu, van~de Weijer, and
  Raducanu]{DBLP:conf/icpr/LiL0R20}
Minghan Li, Xialei Liu, Joost van~de Weijer, and Bogdan Raducanu.
\newblock Learning to rank for active learning: {A} listwise approach.
\newblock In \emph{25th International Conference on Pattern Recognition, {ICPR}
  2020, Virtual Event / Milan, Italy, January 10-15, 2021}, pp.\  5587--5594.
  {IEEE}, 2020.

\bibitem[Li et~al.(2021)Li, Gao, Niu, Xiao, Liu, Liu, Wu, and
  Wang]{DBLP:conf/acl/LiGNXLL0020}
Wei Li, Can Gao, Guocheng Niu, Xinyan Xiao, Hao Liu, Jiachen Liu, Hua Wu, and
  Haifeng Wang.
\newblock {UNIMO:} towards unified-modal understanding and generation via
  cross-modal contrastive learning.
\newblock In \emph{Proceedings of the 59th Annual Meeting of the Association
  for Computational Linguistics and the 11th International Joint Conference on
  Natural Language Processing, {ACL/IJCNLP} 2021, (Volume 1: Long Papers),
  Virtual Event, August 1-6, 2021}, pp.\  2592--2607. Association for
  Computational Linguistics, 2021.

\bibitem[Lin et~al.(2017)Lin, Goyal, Girshick, He, and
  Doll{\'{a}}r]{DBLP:conf/iccv/LinGGHD17}
Tsung{-}Yi Lin, Priya Goyal, Ross~B. Girshick, Kaiming He, and Piotr
  Doll{\'{a}}r.
\newblock Focal loss for dense object detection.
\newblock In \emph{{IEEE} International Conference on Computer Vision, {ICCV}
  2017, Venice, Italy, October 22-29, 2017}, pp.\  2999--3007. {IEEE} Computer
  Society, 2017.

\bibitem[Liu et~al.(2019)Liu, Ott, Goyal, Du, Joshi, Chen, Levy, Lewis,
  Zettlemoyer, and Stoyanov]{DBLP:journals/corr/abs-1907-11692}
Yinhan Liu, Myle Ott, Naman Goyal, Jingfei Du, Mandar Joshi, Danqi Chen, Omer
  Levy, Mike Lewis, Luke Zettlemoyer, and Veselin Stoyanov.
\newblock Roberta: {A} robustly optimized {BERT} pretraining approach.
\newblock \emph{CoRR}, abs/1907.11692, 2019.

\bibitem[Ma et~al.(2021)Ma, Ilievski, Francis, Bisk, Nyberg, and
  Oltramari]{DBLP:conf/aaai/MaIFBNO21}
Kaixin Ma, Filip Ilievski, Jonathan Francis, Yonatan Bisk, Eric Nyberg, and
  Alessandro Oltramari.
\newblock Knowledge-driven data construction for zero-shot evaluation in
  commonsense question answering.
\newblock In \emph{Thirty-Fifth {AAAI} Conference on Artificial Intelligence,
  {AAAI} 2021, Thirty-Third Conference on Innovative Applications of Artificial
  Intelligence, {IAAI} 2021, The Eleventh Symposium on Educational Advances in
  Artificial Intelligence, {EAAI} 2021, Virtual Event, February 2-9, 2021},
  pp.\  13507--13515. {AAAI} Press, 2021.

\bibitem[MacCartney \& Manning(2007)MacCartney and
  Manning]{DBLP:conf/acl/MacCartneyM07}
Bill MacCartney and Christopher~D. Manning.
\newblock Natural logic for textual inference.
\newblock In \emph{Proceedings of the ACL-PASCAL@ACL 2007 Workshop on Textual
  Entailment and Paraphrasing, Prague, Czech Republic, June 28-29, 2007}, pp.\
  193--200. Association for Computational Linguistics, 2007.

\bibitem[Mikolov et~al.(2013)Mikolov, Chen, Corrado, and
  Dean]{DBLP:journals/corr/abs-1301-3781}
Tom{\'{a}}s Mikolov, Kai Chen, Greg Corrado, and Jeffrey Dean.
\newblock Efficient estimation of word representations in vector space.
\newblock In \emph{1st International Conference on Learning Representations,
  {ICLR} 2013, Scottsdale, Arizona, USA, May 2-4, 2013, Workshop Track
  Proceedings}, 2013.

\bibitem[Mostafazadeh et~al.(2016)Mostafazadeh, Chambers, He, Parikh, Batra,
  Vanderwende, Kohli, and Allen]{DBLP:conf/naacl/MostafazadehCHP16}
Nasrin Mostafazadeh, Nathanael Chambers, Xiaodong He, Devi Parikh, Dhruv Batra,
  Lucy Vanderwende, Pushmeet Kohli, and James~F. Allen.
\newblock A corpus and cloze evaluation for deeper understanding of commonsense
  stories.
\newblock In \emph{{NAACL} {HLT} 2016, The 2016 Conference of the North
  American Chapter of the Association for Computational Linguistics: Human
  Language Technologies, San Diego California, USA, June 12-17, 2016}, pp.\
  839--849. The Association for Computational Linguistics, 2016.

\bibitem[Paul \& Frank(2020)Paul and Frank]{DBLP:conf/emnlp/PaulF20}
Debjit Paul and Anette Frank.
\newblock Social commonsense reasoning with multi-head knowledge attention.
\newblock In \emph{Findings of the Association for Computational Linguistics:
  {EMNLP} 2020, Online Event, 16-20 November 2020}, volume {EMNLP} 2020 of
  \emph{Findings of {ACL}}, pp.\  2969--2980. Association for Computational
  Linguistics, 2020.

\bibitem[Pearl(1986)]{pearl1986evidential}
Judea Pearl.
\newblock On evidential reasoning in a hierarchy of hypotheses.
\newblock \emph{Artificial Intelligence}, 28\penalty0 (1):\penalty0 9--15,
  1986.

\bibitem[Pennington et~al.(2014)Pennington, Socher, and
  Manning]{DBLP:conf/emnlp/PenningtonSM14}
Jeffrey Pennington, Richard Socher, and Christopher~D. Manning.
\newblock Glove: Global vectors for word representation.
\newblock In \emph{Proceedings of the 2014 Conference on Empirical Methods in
  Natural Language Processing, {EMNLP} 2014, October 25-29, 2014, Doha, Qatar,
  {A} meeting of SIGDAT, a Special Interest Group of the {ACL}}, pp.\
  1532--1543. {ACL}, 2014.

\bibitem[Peters et~al.(2018)Peters, Neumann, Iyyer, Gardner, Clark, Lee, and
  Zettlemoyer]{DBLP:conf/naacl/PetersNIGCLZ18}
Matthew~E. Peters, Mark Neumann, Mohit Iyyer, Matt Gardner, Christopher Clark,
  Kenton Lee, and Luke Zettlemoyer.
\newblock Deep contextualized word representations.
\newblock In \emph{Proceedings of the 2018 Conference of the North American
  Chapter of the Association for Computational Linguistics: Human Language
  Technologies, {NAACL-HLT} 2018, New Orleans, Louisiana, USA, June 1-6, 2018,
  Volume 1 (Long Papers)}, pp.\  2227--2237. Association for Computational
  Linguistics, 2018.

\bibitem[Qin et~al.(2010)Qin, Liu, and Li]{DBLP:journals/ir/QinLL10}
Tao Qin, Tie{-}Yan Liu, and Hang Li.
\newblock A general approximation framework for direct optimization of
  information retrieval measures.
\newblock \emph{Inf. Retr.}, 13\penalty0 (4):\penalty0 375--397, 2010.

\bibitem[Radford et~al.(2018)Radford, Narasimhan, Salimans, and
  Sutskever]{radford2018improving}
Alec Radford, Karthik Narasimhan, Tim Salimans, and Ilya Sutskever.
\newblock Improving language understanding with unsupervised learning.
\newblock 2018.

\bibitem[Williams et~al.(2018)Williams, Nangia, and
  Bowman]{DBLP:conf/naacl/WilliamsNB18}
Adina Williams, Nikita Nangia, and Samuel~R. Bowman.
\newblock A broad-coverage challenge corpus for sentence understanding through
  inference.
\newblock In \emph{Proceedings of the 2018 Conference of the North American
  Chapter of the Association for Computational Linguistics: Human Language
  Technologies, {NAACL-HLT} 2018, New Orleans, Louisiana, USA, June 1-6, 2018,
  Volume 1 (Long Papers)}, pp.\  1112--1122. Association for Computational
  Linguistics, 2018.

\bibitem[Yu et~al.(2020)Yu, Zhang, Song, Ng, and
  Shang]{DBLP:conf/akbc/YuZSNS20}
Changlong Yu, Hongming Zhang, Yangqiu Song, Wilfred Ng, and Lifeng Shang.
\newblock Enriching large-scale eventuality knowledge graph with entailment
  relations.
\newblock In \emph{Conference on Automated Knowledge Base Construction, {AKBC}
  2020, Virtual, June 22-24, 2020}, 2020.

\bibitem[Zellers et~al.(2018)Zellers, Bisk, Schwartz, and
  Choi]{DBLP:conf/emnlp/ZellersBSC18}
Rowan Zellers, Yonatan Bisk, Roy Schwartz, and Yejin Choi.
\newblock {SWAG:} {A} large-scale adversarial dataset for grounded commonsense
  inference.
\newblock In \emph{Proceedings of the 2018 Conference on Empirical Methods in
  Natural Language Processing, Brussels, Belgium, October 31 - November 4,
  2018}, pp.\  93--104. Association for Computational Linguistics, 2018.

\bibitem[Zhou et~al.(2021)Zhou, Lee, Selvam, Lee, and
  Ren]{DBLP:conf/iclr/ZhouLSL021}
Wangchunshu Zhou, Dong{-}Ho Lee, Ravi~Kiran Selvam, Seyeon Lee, and Xiang Ren.
\newblock Pre-training text-to-text transformers for concept-centric common
  sense.
\newblock In \emph{9th International Conference on Learning Representations,
  {ICLR} 2021, Virtual Event, Austria, May 3-7, 2021}. OpenReview.net, 2021.

\bibitem[Zhu et~al.(2020)Zhu, Pang, Lan, and Cheng]{DBLP:conf/sigir/ZhuPLC20}
Yunchang Zhu, Liang Pang, Yanyan Lan, and Xueqi Cheng.
\newblock L2r{\({^2}\)}: Leveraging ranking for abductive reasoning.
\newblock In \emph{Proceedings of the 43rd International {ACM} {SIGIR}
  conference on research and development in Information Retrieval, {SIGIR}
  2020, Virtual Event, China, July 25-30, 2020}, pp.\  1961--1964. {ACM}, 2020.

\end{thebibliography}
